\documentclass[11pt]{article}

\usepackage[margin=1in]{geometry}
\usepackage[T1]{fontenc}
\usepackage[utf8]{inputenc}
\usepackage{lmodern}
\usepackage{microtype}
\usepackage{amsmath,amssymb,amsfonts}
\usepackage{graphicx}
\usepackage{float}
\usepackage{booktabs}
\usepackage{tabularx}
\usepackage{longtable}
\usepackage{array}
\usepackage{xspace}
\usepackage{xcolor}
\usepackage{enumitem}
\usepackage{listings}
\usepackage{tikz}
\usetikzlibrary{arrows.meta,positioning,fit,calc,shapes.geometric}
\usepackage[colorlinks=true,linkcolor=blue,citecolor=blue,urlcolor=blue]{hyperref}
\usepackage[nameinlink]{cleveref}

\setlist[itemize]{leftmargin=*,topsep=2pt,itemsep=2pt}
\setlist[enumerate]{leftmargin=*,topsep=2pt,itemsep=2pt}
\AddToHook{shipout/foreground}{%
  \begin{tikzpicture}[remember picture,overlay]
    \node[anchor=north west,inner sep=0pt]
      at ([xshift=1in,yshift=-0.42in]current page.north west)
      {\includegraphics[height=14pt]{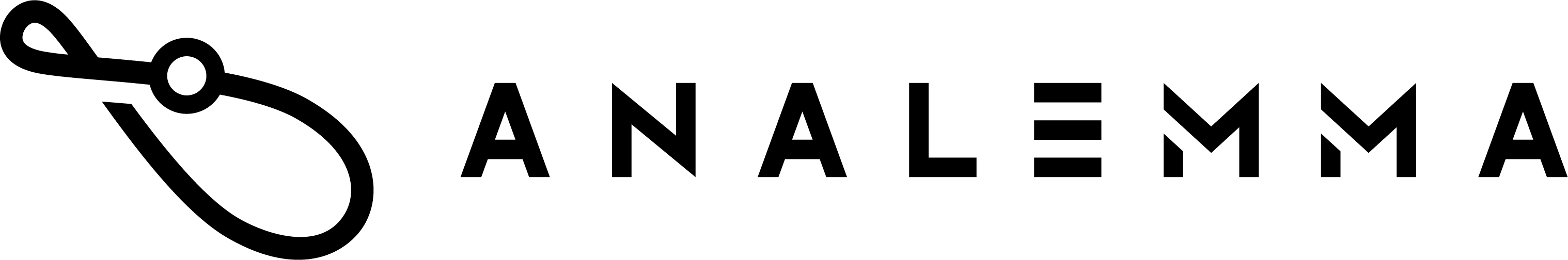}};
    \draw[line width=0.4pt]
      ([xshift=1in,yshift=-0.70in]current page.north west) --
      ([xshift=-1in,yshift=-0.70in]current page.north east);
  \end{tikzpicture}%
}

\newcommand{\ca}{\texttt{claim\_audit.md}\xspace}
\newcommand{\syn}{\texttt{synthesis.md}\xspace}
\newcommand{\ideas}{\texttt{idea\_candidates.md}\xspace}
\newcommand{\rp}{\texttt{research\_proposal.md}\xspace}

\title{\textbf{IdeaTrail: Full-Process Agent Trajectories for Scientific Ideation}}

\author{
  Hengquan Guo\thanks{This work was done while Hengquan Guo was a research intern at Analemma.} \\
  ShanghaiTech University \\
  \texttt{guohq@shanghaitech.edu.cn}
}

\date{}

\begin{document}
\maketitle

\begin{abstract}
% [original]
% Scientific research is a complex, multi-stage workflow rather than a single act of text generation. A research result typically emerges through literature search, paper reading, tool use, claim checking, cross-paper synthesis, brainstorming, rejection of weak directions, and iterative writing. However, existing scientific data resources largely expose final artifacts such as papers or proposals, while high-quality trajectories describing how such artifacts are produced remain scarce.

% [revised]
Scientific ideation unfolds over multiple stages, including literature search, paper reading, tool use, claim checking, cross-paper synthesis, brainstorming, rejection of weak directions, and iterative writing. Yet most existing resources capture isolated components or final artifacts rather than the process connecting them. We introduce IdeaTrail, a dataset of 1,170 multi-turn trajectories for scientific ideation and proposal generation. Each trajectory follows a research process from evidence gathering to either idea selection or proposal construction, jointly recording tool use, acquired evidence, intermediate artifacts, and reasoning. IdeaTrail is synthesized from human-selected research papers and proposal artifacts through a Generator--Advisor loop. The Generator produces the visible sequence of actions, observations, and artifact edits, while the Advisor uses the full generation context to check grounding, causal order, naturalness, and leakage from hidden targets. This reverse-to-forward design keeps trajectories aligned with real scientific artifacts while retaining the uncertainty, evidence use, and staged convergence characteristic of research practice. IdeaTrail provides both reusable process supervision and a general recipe for constructing scientific-research-agent data.
\end{abstract}

\begin{center}
\textbf{Dataset:} \href{https://huggingface.co/datasets/AliceKJ/IdeaTrail}{\texttt{IdeaTrail}}
\end{center}

\section{Introduction}

% [revised]
Scientific ideation is a long-horizon workflow rather than a single generation step. Ideas and proposals emerge through repeated literature search, paper reading, evidence checking, cross-paper synthesis, brainstorming, idea selection, and writing. Recent AI scientist systems automate increasingly large portions of this workflow, from hypothesis generation to experimentation and manuscript production \cite{lu2024aiscientist,gottweis2025aicoscientist,tang2026fars}. In parallel, datasets and training frameworks formulate scientific ideation as idea evaluation, literature-grounded generation, or decomposed discovery \cite{baek2024researchagent,guo2024ideabench,yang2026moosestar,zhao2026agenticideation}. Public supervision for the full trajectory, however, remains limited: existing resources usually emphasize final ideas, selected decisions, or system outputs rather than the connected sequence of evidence discovery, tool interaction, reasoning, artifact evolution, idea selection, and proposal construction.

This report introduces IdeaTrail, a dataset of reverse-synthesized process supervision for scientific ideation. Each instance records a multi-turn trajectory from a broad research query to either an idea-level or proposal-level endpoint. The message stream preserves tool invocations, evidence-gathering steps, reasoning turns, and artifact updates. In proposal-extension trajectories, the final proposal bridges ideation and downstream implementation by specifying the research question, core mechanism, expected contribution, evaluation setup, and risks in enough detail to support coding, experimentation, and verification.

Trajectories are generated through a Generator--Advisor review loop. The Advisor has access to the full generation context, including the selected paper, proposal endpoint, and hidden constraints. The Generator sees only the visible query, current trajectory prefix, current artifacts, and available tools, from which it proposes forward actions, observations, reasoning steps, and artifact edits. The Advisor then evaluates whether each step is grounded, causally ordered, natural, and free of leakage from hidden targets. This iterative verification constrains drift, fabricated evidence, and premature disclosure while preserving a plausible forward research process.

The Generator's tools are cutoff-aware. Each case is generated under a specified information horizon, preventing the agent from using evidence unavailable at the intended cutoff. This constraint matters because later papers, benchmarks, or terminology can simplify reconstruction while misrepresenting the information state in which the target direction originally emerged. Cutoff-bounded tool use therefore reduces temporal leakage and better aligns the generated process with the target scientific artifact.

IdeaTrail also provides researcher portraits as optional conditioning context. Each portrait is distilled from prior papers by the target artifact's first author and summarizes domain knowledge, research approach, novelty style, evidence preferences, and recurring patterns. It acts as a research prior, encouraging trajectories that reflect plausible researcher-specific choices rather than generic ideation behavior.

In summary, IdeaTrail contributes both a multi-turn trajectory dataset for scientific ideation and proposal generation and a reverse-to-forward recipe for constructing process supervision from existing research artifacts.

\section{Related Work}

\paragraph{The emergence of AI scientists.}
Recent systems suggest that language-model agents can participate in increasingly broad portions of the research lifecycle. The AI Scientist connects ideation with implementation, experimentation, manuscript writing, and automated review \cite{lu2024aiscientist}. Google's Co-Scientist uses a multi-agent process of generation, critique, ranking, and refinement to develop experimentally testable hypotheses \cite{gottweis2025aicoscientist}. More recently, FARS demonstrated a fully automated system operating across ideation, planning, experimentation, and writing at deployment scale, while retaining proposals, code, logs, results, and manuscripts in a shared workspace \cite{tang2026fars}. Together, these systems motivate training resources that represent research as an extended process rather than an isolated generation task.

\paragraph{Abstractions in existing ideation supervision.}
Existing resources make scientific ideation more tractable by concentrating on selected outputs or decomposed decisions. IdeaBench evaluates generated ideas against reference research contributions, while ResearchAgent iteratively generates and reviews literature-grounded ideas \cite{guo2024ideabench,baek2024researchagent}. MOOSE-Star identifies the combinatorial difficulty of directly learning $P(h\mid b)$ and constructs TOMATO-Star by decomposing paper-derived discovery into motivation planning, inspiration retrieval, and incremental hypothesis composition \cite{yang2026moosestar}. This decomposition provides scalable supervision for key discovery decisions, but intentionally abstracts away much of the temporally extended activity through which evidence is searched, read, checked, synthesized, and converted into evolving research artifacts. Agentic-Ideation moves toward process supervision by using oracle guidance to synthesize tool-using ideation trajectories \cite{zhao2026agenticideation}, yet reusable data capturing the broader, persistent research process remains limited.

\paragraph{Positioning IdeaTrail.}
IdeaTrail addresses this gap by representing scientific ideation as a long-horizon, multi-turn trajectory rather than only a final idea or a sequence of decomposed choices. Each trajectory connects literature search and reading, tool observations, explicit evidence artifacts, cross-paper synthesis, candidate comparison, and an idea-level or proposal-level endpoint. Its reverse-to-forward Generator--Advisor construction retains alignment with real research artifacts while exposing the intermediate process as reusable supervision. IdeaTrail is therefore not another autonomous scientist system, but a dataset and synthesis recipe for training and studying agents that must sustain coherent research behavior over many turns.

\section{The Forward Process of IdeaTrail: How Research Ideation Is Generated}

This section defines the forward process of ideation. The process is a design convention rather than a claim about how every human scientist works. Its purpose is to provide a stable training target for research agents. Within this convention, scientific ideation is modeled as a coupled sequence of behaviors and artifacts.

The behavior sequence describes what the agent does. The artifact sequence records what the agent has learned and decided. The two views are inseparable: a search action has limited training value unless it changes what the agent knows; an artifact has limited credibility unless it can be traced back to search, reading, and reasoning actions.

\subsection{Behavioral Trajectory: From Exploration to Proposal Writing}

A trajectory is represented as a sequence of turns:

\begin{equation}
\tau = \left\{x_t\right\}_{t=1}^{T}, \qquad x_t = (b_t, u_t, o_t, \Delta a_t),
\end{equation}

where $b_t$ is the agent behavior at turn $t$, $u_t$ is the tool call or action, $o_t$ is the observation or tool result, and $\Delta a_t$ is the artifact update induced by the turn. The artifact state evolves as

\begin{equation}
 a_t = \textsc{Update}(a_{t-1}, \Delta a_t).
\end{equation}

The agent first gathers evidence through search and reading, then audits claims against sources, synthesizes cross-paper tensions and opportunities, compares candidate ideas, and finally writes an executable proposal. The important property is staged convergence: the final idea should emerge from evidence and alternatives rather than appear as a known answer from the first turn.

\begin{figure}[t]
\centering
\includegraphics[width=\textwidth]{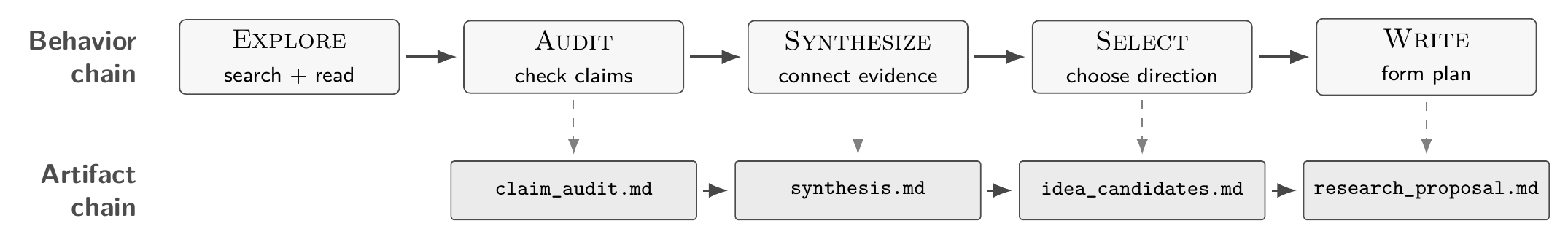}
\caption{Canonical forward process. The top row shows behavioral stages. The bottom row shows durable artifacts. The training target is the coupled evolution of both rows.}
\label{fig:forward-process}
\end{figure}

\subsection{Artifact Trajectory}

The artifact trajectory is the durable record of the agent's research state:
\begin{equation}
\ca \rightarrow \syn \rightarrow \ideas \rightarrow \rp.
\end{equation}
Each artifact has a distinct role. \ca is the evidence ledger, grounding claims in sources and recording limitations and implications. \syn compresses audited evidence into cross-paper tensions, mechanisms, boundary conditions, and research opportunities. \ideas records candidate directions, rejected alternatives, and the selection logic behind the winning idea. \rp expands that idea into an executable proposal with motivation, method, evaluation, risks, and implementation details.

\section{IdeaTrail Construction: Reverse-Synthesized Research Trajectories}

% [original]
% This section describes how \method constructs research trajectories. As shown in \Cref{fig:datagen}, the pipeline first derives Advisor-only constraints and anchor artifacts from human-selected final research artifacts, then asks the Generator to produce a chronological trajectory with tool calls, observations, and artifact edits. The Advisor verifies that the generated process remains grounded, causally ordered, natural, and free of leakage from hidden targets.

\begin{figure}[t]
\centering
\includegraphics[width=\textwidth]{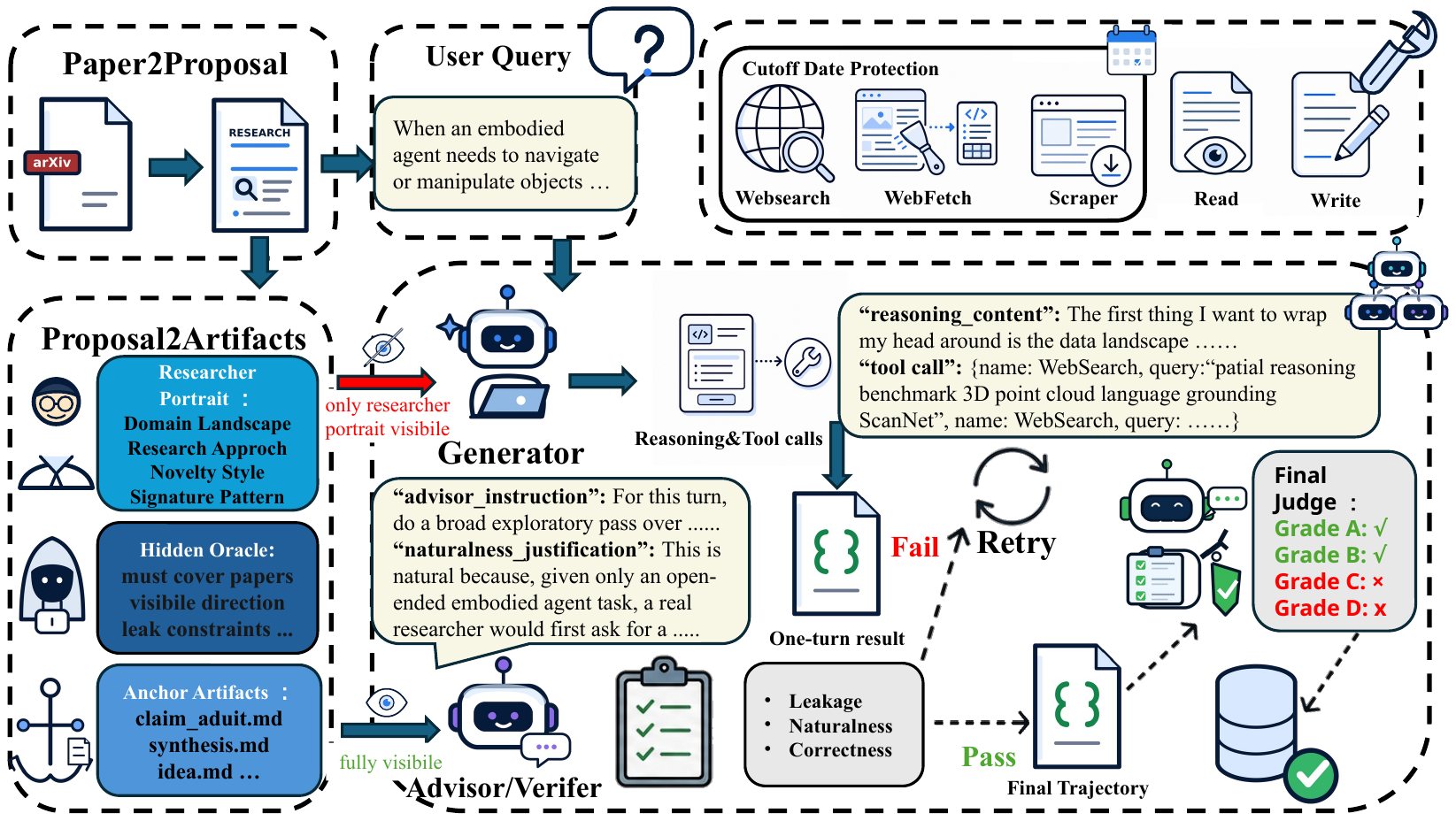}
\caption{Overview of the IdeaTrail data-generation pipeline.}
\label{fig:datagen}
\end{figure}

% [revised]
This section describes how IdeaTrail constructs research trajectories. As shown in \Cref{fig:datagen}, the pipeline derives Advisor-only constraints and anchor artifacts from human-selected final research artifacts, after which the Generator produces a chronological sequence of tool calls, observations, reasoning steps, and artifact edits. The Advisor verifies that this sequence remains grounded, causally ordered, natural, and free of leakage from hidden targets.

\subsection{The Hindsight Problem}

Let $p$ denote a final research proposal and let $\tau$ denote a plausible forward trajectory that could have produced it. A naive inverse-generation setup would ask a Generator to solve

\begin{equation}
 p \rightarrow \tau.
\end{equation}

This inverse problem is underdetermined: many trajectories can lead to the same final proposal. Directly conditioning the Generator on that proposal may produce superficially plausible early turns that nevertheless depend on the final method, benchmark, terminology, or decision rule. Such dependence introduces target leakage, unnatural search behavior, and overly linear reasoning. IdeaTrail instead separates Advisor-side constraint construction from Generator-side trajectory synthesis:

\begin{equation}
 p \rightarrow o_{\mathrm{adv}} \rightarrow a^{\star} \rightarrow \hat{\tau}_{\mathrm{gen}},
\end{equation}

where $o_{\mathrm{adv}}$ is Advisor-only oracle context, $a^{\star}$ is a set of anchor artifacts, and $\hat{\tau}_{\mathrm{gen}}$ is the generated trajectory. The Advisor uses the oracle context and anchors to constrain and verify the target, while the Generator produces the visible trajectory in chronological order.

\subsection{Advisor Context, Anchors, and Trajectories}

The Advisor-only context is extracted from the final proposal. It includes final atoms, artifact atoms, causal dependencies, and leakage locks. Final atoms summarize the thesis, method, evaluation, decision rule, novelty, risks, and other proposal-critical components. Artifact atoms specify what the intermediate artifacts should eventually discover. Leakage locks specify which final atoms may not appear before a given stage.

Anchor artifacts are deterministic targets built from the proposal seed and Advisor-only context. They include anchor versions of the claim audit, synthesis, idea candidates, and proposal. These anchors are not shown to the Generator as trajectory inputs; they are used by the Advisor and verifiers as target constraints. The generated trajectory is still allowed to contain local uncertainty, detours, rejected ideas, and incremental edits.

\subsection{Synthesis Objective}

The generated trajectory should satisfy four constraints:

\begin{align}
\text{Convergence}(\hat{\tau}_{\mathrm{gen}}, a^{\star}) &\geq \gamma, \\
\text{Leakage}(\hat{\tau}_{\mathrm{gen}}, o_{\mathrm{adv}}) &= 0, \\
\text{Grounding}(\hat{\tau}_{\mathrm{gen}}) &\geq \eta, \\
\text{Naturalness}(\hat{\tau}_{\mathrm{gen}}) &\geq \nu.
\end{align}

Convergence means the final idea or proposal aligns with the anchor's core mechanism. Leakage means Advisor-only terms or final conclusions do not appear too early. Grounding means claims are supported by observed sources. Naturalness means the trajectory contains realistic uncertainty, reading, revision, and staged convergence.

\subsection{Proposal Ingestion}

The pipeline begins with \rp. The ingestion step parses the proposal into a structured proposal seed. This step is conservative: it extracts information already present in the proposal and avoids adding missing details. The proposal seed is the factual base for downstream reverse planning. A simplified seed schema is shown below.

\begin{lstlisting}
{
  "proposal_id": "case_id",
  "research_question": "...",
  "thesis": "...",
  "method": "...",
  "evaluation": "...",
  "baselines": [...],
  "risks": [...],
  "novelty_claim": "...",
  "parse_quality": "high | medium | low"
}
\end{lstlisting}

\subsection{Visible Direction Generation}

The visible direction converts the initial research question and allowed context into a natural user message. It gives the agent enough information to begin research, but it excludes proposal-specific method names, final titles, benchmark names, and internal pipeline terms. An exact-term leakage check is applied before the visible direction is accepted.

\subsection{Anchor Artifact Construction}

The anchor builder creates four target artifacts:

\begin{equation}
 a^{\star} = (a^{\star}_{\text{claim}}, a^{\star}_{\text{synth}}, a^{\star}_{\text{idea}}, a^{\star}_{\text{proposal}}).
\end{equation}

The anchor claim audit is grounded in searched or parsed papers. The anchor synthesis expands the oracle's synthesis atoms into problem levels, evidence base, mechanism hypotheses, contradictions, boundaries, and candidate directions. The anchor idea artifact records the winning idea. The anchor proposal is the most faithful target version of the original proposal.

\subsection{High-Level Milestone Planning}

Given the hidden oracle, anchors, and a turn budget, the high-level planner creates milestones. A milestone specifies what should have been achieved by a certain point in the trajectory. For example, a 50-turn trajectory may allocate early turns to broad evidence gathering, middle turns to claim audit and synthesis, and late turns to idea selection and proposal writing.

\subsection{Turn Generation and Public Serialization}

% [original]
% The trajectory generator produces one turn at a time. A turn may include an assistant message, tool call, tool result summary, and artifact update. The output is stored in a turn-level JSONL format suitable for supervised fine-tuning.
%
% \begin{lstlisting}
% {
%   "case_id": "case_id",
%   "turn_id": 14,
%   "stage": "S1_CLAIM_AUDIT",
%   "milestone_id": "M2",
%   "assistant_message": "I have enough raw evidence to start the claim audit.",
%   "tool_call": {
%     "name": "Write",
%     "input_summary": "Create claim_audit.md with the first audited claim."
%   },
%   "tool_result": {
%     "status": "success",
%     "summary": "claim_audit.md written"
%   },
%   "artifact_updates": [
%     {"artifact": "claim_audit.md", "operation": "write"}
%   ]
% }
% \end{lstlisting}

% [revised]
The trajectory generator produces one turn at a time, but the public release
does not serialize turns as separate JSONL records. Instead, each JSONL line
contains one complete trajectory. Turn-level assistant messages, tool calls,
tool results, and artifact-edit actions are retained inside the ordered
\texttt{messages} array. The top-level release schema is illustrated below;
underscore-prefixed fields are release metadata and may be absent.

\begin{lstlisting}
{
  "sample_id": "...",
  "messages": [...],
  "tools": [...],
  "parallel_tool_calls": true,
  "_src": "...",
  "_grade": "A | B | C | None",
  "_naturalness": ...,
  "_proposal_synth": true,
  "_proposal_path": "..."
}
\end{lstlisting}

This schema is the public trajectory-level representation. Internal
synthesis-time state may additionally track turn identifiers, stages,
milestones, or artifact deltas, but those fields do not constitute the schema
of the public release. In particular, \texttt{\_grade} may also be missing,
and the literal string \texttt{"None"} is distinct from a missing key.

\subsection{Verifier-Guided Retry}

The verifier checks each turn or milestone segment. It does not generate new trajectory content. It returns pass or retry instructions. Typical retry triggers include forbidden-term leakage, creating an artifact too early, using a paper title before it appears in search observations, fabricating a URL, or drifting away from the target problem space.

\section{Dataset Format}

\subsection{Trajectory Types}

Not every IdeaTrail trajectory extends to a research proposal. After the Generator produces \ideas, an Advisor-side judge checks whether the selected idea matches the ground-truth target in its core problem, mechanism, and research direction. Because the synthesis process permits local uncertainty and exploration, some useful ideas diverge slightly from the target. These cases are retained as idea-level trajectories rather than expanded into \rp.

% [original]
% The dataset therefore records two endpoint types. In \textit{idea-only} cases, the system prompt guides the Generator only through evidence gathering, claim audit, synthesis, and idea candidate selection. In \textit{proposal-extension} cases, the idea passes the convergence judge and the prompt continues to proposal writing. This distinction prevents weakly aligned ideas from being forced into ground-truth proposals while still preserving useful ideation supervision.

% [revised]
The dataset therefore records two endpoint types. In \textit{idea-only} cases, the system prompt guides the Generator only through evidence gathering, claim audit, synthesis, and idea candidate selection. In \textit{proposal-extension} cases, the idea passes the convergence judge and the prompt continues to proposal writing. This distinction prevents weakly aligned ideas from being forced into ground-truth proposals while still preserving useful ideation supervision. The release contains 688 proposal-extension trajectories, marked by \texttt{\_proposal\_synth=true}; these records also contain \texttt{\_proposal\_path}. The remaining 482 idea-level trajectories omit both fields. Thus, endpoint type uses a true-versus-absent encoding rather than a conventional Boolean true/false field.

\subsection{Value-Tier Labels}

% [original]
% The dataset includes value-tier labels to support weighted training. The current trajectories have not yet undergone aggressive post-hoc filtering, and many turns contain long reasoning traces or other low-information intermediate text. Treating every turn as equally valuable could over-train agents on redundant reasoning styles and dilute the supervision from turns that actually drive evidence use, idea selection, and proposal construction. The label is therefore not merely a measure of surface text quality; it estimates each turn's contribution to the final idea or proposal.

% [revised]
The dataset includes value-tier labels to support weighted training. Because the release preserves long reasoning traces and low-information intermediate text, weighting all turns equally can overemphasize redundant reasoning and dilute supervision from decisions that advance evidence use, idea selection, or proposal construction. The labels therefore estimate each turn's contribution to the final idea or proposal rather than its surface-level writing quality.

\begin{itemize}
  \item \textbf{High:} turns that diagnose root causes, introduce nontrivial mechanisms, select or revise the winning idea, formalize the proposal method, or fix final proposal quality.
  \item \textbf{Normal:} useful research work such as reading, ordinary claim auditing, artifact drafting, and evidence consolidation.
  \item \textbf{Low:} mechanical, redundant, shallow, or formatting-heavy turns.
\end{itemize}

\section{Quality Control Protocol}

\subsection{One-Vote Failure Conditions}

A trajectory is rejected or routed to repair if it triggers any one-vote failure condition:

\begin{itemize}
  \item \textbf{Oracle leakage:} internal terms or final conclusions appear too early.
  \item \textbf{Title pre-knowledge:} a complete paper title is searched before it was observed.
  \item \textbf{Fabricated evidence:} a claim, result, citation, or paper conclusion is invented or misattributed.
  \item \textbf{Fake URL:} a scraper URL is not derived from a prior search observation.
  \item \textbf{Target drift:} the final idea or proposal moves outside the intended problem space or core mechanism.
\end{itemize}

\subsection{Trajectory Rubric}

The main trajectory rubric uses the following five dimensions.

\begin{table}[H]
\centering
\small
\begin{tabularx}{\textwidth}{p{2.6cm}X}
\toprule
Dimension & Question \\
\midrule
Safety & Does the trajectory avoid leakage, hallucination, fake URLs, and unsupported paper titles? \\
Naturalness & Does the reasoning look like progressive research work with uncertainty, revision, and staged convergence? \\
Completeness & Are important papers deeply read and incorporated into \ca and \syn? \\
Convergence & Does the final idea align with the anchor's problem space and mechanism while remaining independently derived? \\
Portrait utility & If a researcher portrait is used, does it meaningfully affect search direction, reasoning style, or final idea? \\
\bottomrule
\end{tabularx}
\caption{Trajectory quality rubric.}
\label{tab:rubric}
\end{table}

\subsection{Artifact Metrics}

Artifact-level metrics include grounding rate, traceability gap rate, proposal recoverability, leakage score, and artifact incrementality. Artifact incrementality measures whether files emerge through skeleton, fill, revise, and finalize operations rather than being written in one unrealistic block.

\section{Dataset Statistics}
\label{sec:dataset-statistics}

\subsection{Counting Protocol}

IdeaTrail contains $N=1{,}170$ long-horizon trajectories for scientific
ideation and research-proposal generation. Each JSONL record has a unique
\texttt{sample\_id}, a multi-turn \texttt{messages} sequence, seven tool
definitions, and release metadata such as \texttt{\_src} and
\texttt{\_grade}.

Three units are kept separate throughout this section: a trajectory is one
JSONL record, a message is one role-tagged conversational event, and a tool
call is one function invocation emitted inside an assistant message. A single
assistant message may therefore contribute several tool calls.

For trajectory $i$, let $m_i$ be its number of messages,
$t_i^{\mathrm{text}}$ its message-text token count, and
$t_i^{\mathrm{json}}$ its full-record token count. Means are computed as
\begin{equation}
\bar{x}=\frac{1}{N}\sum_{i=1}^{N}x_i,
\qquad x_i\in\{m_i,t_i^{\mathrm{text}},t_i^{\mathrm{json}}\}.
\end{equation}

Text-field counts concatenate all string-valued \texttt{content} and
\texttt{reasoning\_content} fields in message order with one newline between
fields. Full-record counts encode the exact decompressed JSONL line without
its trailing newline.

Both views use the \texttt{o200k\_base} tokenizer. Strings resembling special
tokens are encoded as ordinary text. Chat templates, packing, and added
control tokens can increase the effective training length beyond these
corpus-level counts.

\subsection{Scale and Integrity}

All 1,170 records parse successfully and all identifiers are unique. Each
trajectory begins with a system message followed by its only user task. The
parallel-call flag is true in every record, and all records expose the same
seven tool definitions.

\begin{table}[t]
\centering
\small
\setlength{\tabcolsep}{4pt}
\begin{tabular}{lr}
\toprule
Statistic & Value \\
\midrule
Trajectories / unique IDs & 1,170 / 1,170 \\
JSON parse errors / duplicate IDs & 0 / 0 \\
Messages & 163,057 \\
Assistant messages & 38,563 \\
Tool messages / tool calls & 122,154 / 122,154 \\
Tool definitions per trajectory & 7 \\
Unique / singleton topics & 963 / 840 \\
Decompressed JSONL & 553.5 MB / 527.9 MiB \\
Compressed JSONL shards & 92.2 MB / 87.9 MiB \\
Parquet shards & 122.9 MB / 117.2 MiB \\
\bottomrule
\end{tabular}
\caption{Corpus scale, release size, and integrity checks. Decimal MB and
binary MiB are reported separately.}
\label{tab:dataset-overview}
\end{table}

% [original]
% The case count is moderate, while each case contains a deep research process.
% Consequently, the corpus contributes 92.19M message-text tokens and 135.98M
% serialized-JSON tokens despite having only 1,170 training examples.

% [revised]
Although the release contains 1,170 trajectories, each captures an extended
research process. The corpus therefore comprises 92.19M message-text tokens
and 135.98M serialized-JSON tokens.

\subsection{Long-Horizon Context}

\begin{figure*}[t]
\centering
\includegraphics[width=\textwidth]{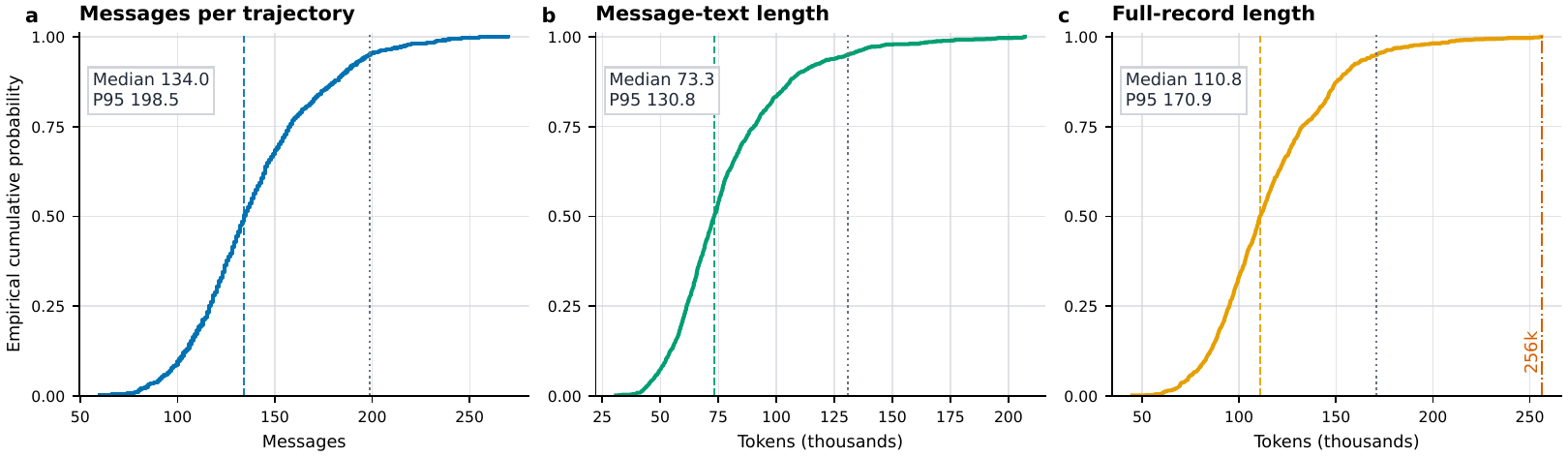}
\caption{Empirical cumulative distributions of trajectory depth. Dashed
colored lines mark medians, dotted gray lines mark P95, and the red reference
in panel (c) marks 256k tokens.}
\label{fig:trajectory-lengths}
\end{figure*}

Table~\ref{tab:length-distribution} reports exact distribution summaries, and
Figure~\ref{fig:trajectory-lengths} shows their empirical cumulative
distributions. Percentiles use NumPy's linear interpolation rule.

\begin{table*}[t]
\centering
\scriptsize
\setlength{\tabcolsep}{4.2pt}
\begin{tabular}{lrrrrrrr}
\toprule
Metric & Min & Median & Mean & P95 & P99 & Max & Total \\
\midrule
Messages / trajectory
  & 60 & 134 & 139.4 & 198.5 & 233.9 & 270 & 163,057 \\
Assistant messages / trajectory
  & 14 & 32 & 33.0 & 46.0 & 51.0 & 63 & 38,563 \\
Tool messages / trajectory
  & 38 & 100 & 104.4 & 154.0 & 185.0 & 217 & 122,154 \\
Text-field tokens / trajectory
  & 30,945 & 73,305 & 78,792 & 130,768 & 175,552 & 207,064 & 92.19M \\
Serialized-JSON tokens / trajectory
  & 45,010 & 110,815 & 116,226 & 170,876 & 212,197 & 255,865 & 135.98M \\
\bottomrule
\end{tabular}
\caption{Trajectory-length distributions under message-only and full-record
token accounting. The serialized view includes keys, metadata, tool schemas,
arguments, and punctuation.}
\label{tab:length-distribution}
\end{table*}

Only 17 trajectories (1.5\%) fit within 64k tokens under full-record
accounting. In contrast, 1,153 trajectories (98.5\%) exceed 64k, 347 (29.7\%)
exceed 128k, and 27 (2.3\%) exceed 192k.

The maximum serialized record reaches 255,865 tokens. This upper tail makes
truncation, sequence packing, attention memory, and loss masking central
design choices rather than incidental preprocessing details.

Text-only training sees a shorter distribution: 351 trajectories (30.0\%)
fit within 64k text tokens, while 66 (5.6\%) exceed 128k. Reporting only the
text view therefore understates the context required by schema-preserving
agent training.

\subsection{Message and Text Composition}

Tool observations dominate both event frequency and text mass. Tool messages
contribute 122,154 of 163,057 messages (74.9\%), followed by assistant messages
at 23.7\%; system and user messages each contribute 0.7\%.

\begin{figure*}[t]
\centering
\includegraphics[width=0.92\textwidth]{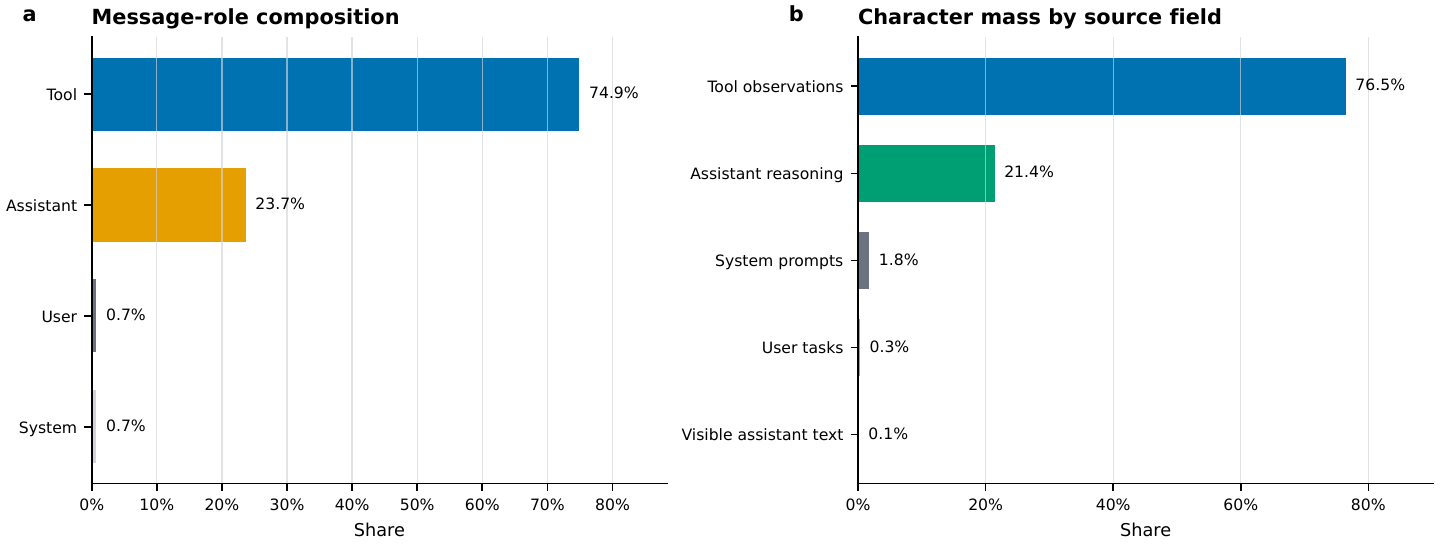}
\caption{Corpus composition by (a) message role and (b) character source.
Shares in panel (b) are computed over all string-valued message text fields.}
\label{fig:corpus-composition}
\end{figure*}

At the character level, tool observations contribute 293.43M characters
(76.5\%). Assistant reasoning contributes another 82.11M (21.4\%). Together
they account for 97.8\% of the recorded text.

Visible assistant content contributes only 0.42M characters (0.1\%). The
primary supervision signal is therefore carried by evidence, intermediate
reasoning, and state-changing actions rather than by short final responses.

\subsection{Tool-Use Structure}

For tool $f$, its call share is
\begin{equation}
p_f=\frac{n_f}{\sum_{f'} n_{f'}},
\end{equation}
where $n_f$ is the number of calls to $f$. Reading consists of \texttt{View};
retrieval combines \texttt{WebSearch} and \texttt{Scraper}; authoring combines
\texttt{Write} and \texttt{Edit}; local discovery combines \texttt{Glob} and
\texttt{Grep}.

\begin{figure*}[t]
\centering
\includegraphics[width=0.96\textwidth]{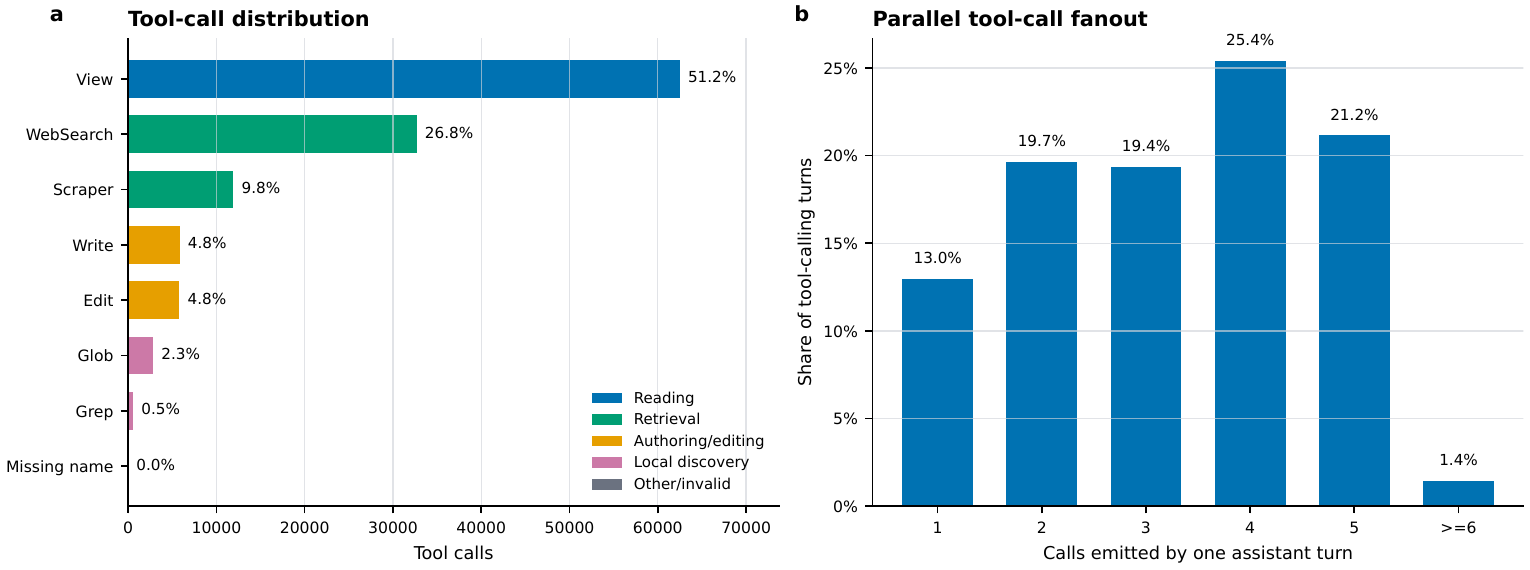}
\caption{Agent tool behavior. Panel (a) reports call volume with tool-family
colors. Panel (b) reports the number of calls emitted by each tool-calling
assistant message.}
\label{fig:tool-use}
\end{figure*}

\begin{table*}[t]
\centering
\scriptsize
\setlength{\tabcolsep}{5pt}
\begin{tabular}{llrrrr}
\toprule
Tool & Family & Calls & Share & Median obs. tokens & Mean obs. chars \\
\midrule
\texttt{View}      & Reading             & 62,485 & 51.2\% & 658   & 2,890 \\
\texttt{WebSearch} & Retrieval           & 32,687 & 26.8\% & 764   & 2,479 \\
\texttt{Scraper}   & Retrieval           & 11,921 & 9.8\%  & 565   & 2,267 \\
\texttt{Write}     & Authoring/editing   & 5,846  & 4.8\%  & 21    & 91 \\
\texttt{Edit}      & Authoring/editing   & 5,803  & 4.8\%  & 1     & 2 \\
\texttt{Glob}      & Local discovery     & 2,839  & 2.3\%  & 265   & 1,253 \\
\texttt{Grep}      & Local discovery     & 562    & 0.5\%  & 140.5 & 1,220 \\
Missing name       & Other/invalid        & 11     & $<$0.1\% & 4   & 14 \\
\bottomrule
\end{tabular}
\caption{Tool-call frequency and observation length. Each tool observation is
matched to the immediately preceding assistant call batch using
\texttt{tool\_call\_id}.}
\label{tab:tool-call-distribution}
\end{table*}

Reading and retrieval account for 62,485 (51.2\%) and 44,608 (36.5\%) calls,
respectively. Their combined share is 87.7\%, compared with 9.5\% for
authoring/editing and 2.8\% for local discovery.

Among 38,563 assistant messages, 37,361 (96.9\%) emit at least one tool call.
The fanout mean is 3.27 calls, the median is 3, P95 is 5, and the maximum is
13. Most assistant turns therefore execute a small batch of coordinated
actions before receiving observations.

Tool-call IDs have batch-local scope. Reuse occurs in 723 trajectories, with
8,464 call occurrences beyond the first use of an ID and a maximum of 36 uses
for one ID. A trajectory-global ID map can silently attach observations to the
wrong calls and distort per-tool observation statistics.

\subsection{Supervision and Release Metadata}

Every assistant message carries non-empty reasoning text. Only 514 (1.3\%)
have non-empty visible content. Among turn-level tier labels, 17,679 are
normal (45.8\%), 13,109 are high (34.0\%), and 6,202 are low (16.1\%); the
label is absent for 1,573 messages (4.1\%).

Trajectory-level \texttt{\_grade} counts are 754 \texttt{A}, 44 \texttt{B},
31 \texttt{C}, 228 literal strings \texttt{"None"}, and 113 missing keys. The
explicit A/B subset therefore contains 798 trajectories (68.2\%).

The source-grade relation is deterministic in this release:
\texttt{v5} contains 94 A and 8 B trajectories; \texttt{rp23} contains 228
literal \texttt{"None"} and 113 missing grades; \texttt{rp4} contains 660 A,
36 B, and 31 C trajectories. These values are release-pipeline metadata and
do not define a portable ordinal quality scale.

\subsection{Topic and Temporal Coverage}

Topic identifiers are extracted from structured tool paths rooted in the
release topic directory. The 1,170 trajectories cover 963 topics; 840 topics
occur once, and the ten most frequent topics account for only 70 trajectories
(6.0\%).

The topic entropy is 9.745 bits. Normalizing by the maximum entropy for 963
observed categories gives
\begin{equation}
H_{\mathrm{norm}}=\frac{-\sum_j p_j\log_2 p_j}{\log_2 963}=0.983,
\end{equation}
which confirms a broad long tail rather than concentration in a few dominant
topics.

\begin{figure*}[t]
\centering
\includegraphics[width=\textwidth]{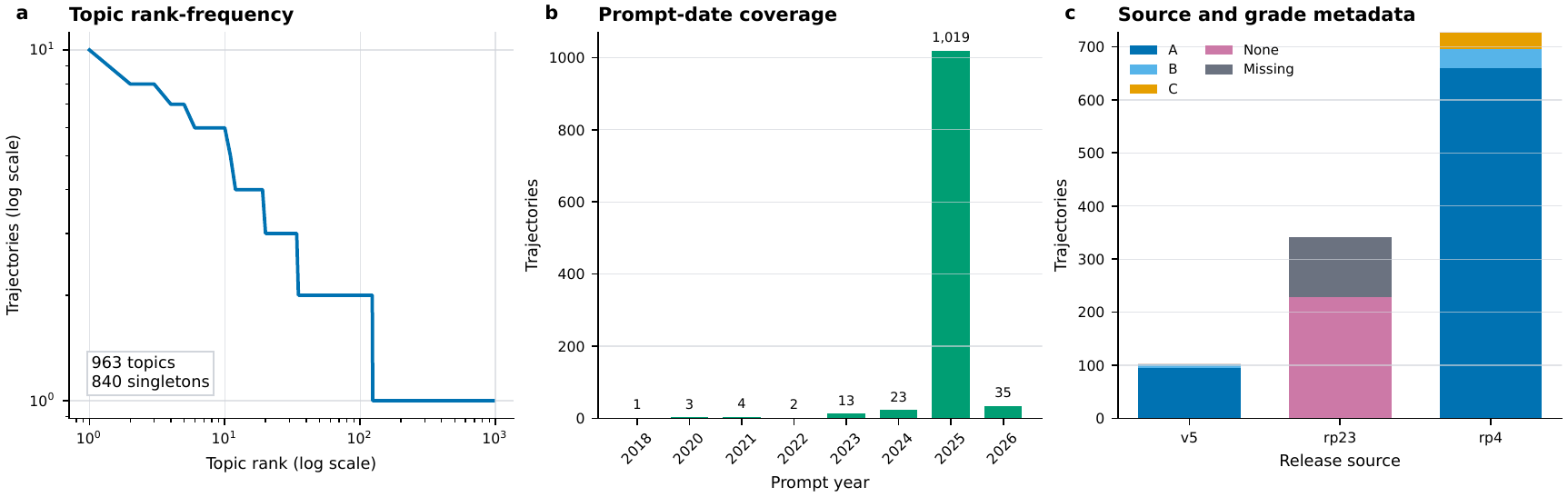}
\caption{Coverage and release metadata. Panel (a) shows the topic
rank-frequency curve, panel (b) summarizes the first valid ISO date in each
system prompt, and panel (c) shows source-conditioned grade composition.}
\label{fig:coverage-metadata}
\end{figure*}

The most frequent topic, single-image 3D reconstruction, appears in 10
trajectories. Two topics have 8 trajectories, two have 7, and five have 6;
the remaining mass is distributed across the long tail.

Prompt dates are parseable for 1,100 trajectories (94.0\%), spanning 196
unique dates from 2018-12-18 to 2026-01-16. Of these, 1,019 fall in 2025.
Prompt dates describe the stated evidence cutoff context and should not be
interpreted as publication or generation timestamps.

Topic and temporal leakage require separate controls. A topic-disjoint split
can still share cutoff periods, while a temporally disjoint split can retain
near-duplicate research themes.

\subsection{Format and Linkage Audit}

\begin{table*}[t]
\centering
\scriptsize
\begin{minipage}[t]{0.43\textwidth}
\centering
\textbf{Source-grade matrix}\\[0.35em]
\setlength{\tabcolsep}{3pt}
\begin{tabular}{lrrrrrr}
\toprule
Source & A & B & C & \texttt{"None"} & Missing & Total \\
\midrule
\texttt{v5}   & 94  & 8  & 0  & 0   & 0   & 102 \\
\texttt{rp23} & 0   & 0  & 0  & 228 & 113 & 341 \\
\texttt{rp4}  & 660 & 36 & 31 & 0   & 0   & 727 \\
\midrule
Total         & 754 & 44 & 31 & 228 & 113 & 1,170 \\
\bottomrule
\end{tabular}
\end{minipage}
\hfill
\begin{minipage}[t]{0.54\textwidth}
\centering
\textbf{Format and linkage checks}\\[0.35em]
\begin{tabular}{lr}
\toprule
Check & Value \\
\midrule
Top-level / message schema variants & 5 / 9 \\
Records with proposal metadata & 688 \\
Records with \texttt{\_reclaim} metadata & 31 \\
Assistant-final / tool-final trajectories & 1,168 / 2 \\
Unmatched tool observations & 0 \\
Calls with empty function name & 11 \\
Trajectories reusing tool-call IDs & 723 \\
Repeated call-ID occurrences & 8,464 \\
Maximum uses of one call ID & 36 \\
Messages with \texttt{\_new=true} & 16,522 \\
Trajectories with special-token-like strings & 42 \\
Unique special-token-like strings & 89 \\
\bottomrule
\end{tabular}
\end{minipage}
\caption{Release metadata and format-level audit. Missing \texttt{\_grade}
means the key is absent; literal \texttt{"None"} remains a string.}
\label{tab:metadata-audit}
\end{table*}

The 688 proposal-path values are sanitized relative construction paths and
serve as auxiliary provenance rather than semantic targets. The 31
\texttt{\_reclaim} values and 113 missing-grade records likewise require
explicit missing-value handling.

Two trajectories terminate with a tool message, and 11 tool calls have an
empty function name. Training code that assumes assistant-final samples or
validates only null tool names will mishandle these cases.

Overall, IdeaTrail is characterized by three coupled properties: long context,
dense tool interaction, and broad topic coverage. A median trajectory contains
134 messages and 110,815 serialized tokens; 87.7\% of calls retrieve or inspect
evidence; and 87.2\% of topic identifiers occur exactly once.

These properties favor training and evaluation protocols that preserve tool
linkage, account for schema tokens, and control topic and temporal leakage.
They also make aggregate sample count alone a poor proxy for the corpus's
computational and supervisory scale.

\section{Limitations}

This release should be viewed as an initial and necessarily incomplete corpus
rather than a fully cleaned training set. Generating and verifying long-horizon
research trajectories is computationally and financially expensive, which
limits both the number of trajectories and the amount of post-processing that
can be applied. The current data have not undergone systematic turn-level
filtering beyond the existing quality-control pipeline. Consequently, some
turns---including turns labeled with a low value tier---still contain lengthy
reasoning, repeated planning, mechanical tool interactions, or other
low-information text. Value-tier labels make such content easier to filter or
down-weight, but they do not remove the noise, and the labels themselves may be
imperfect.

The corpus also has limited process diversity. Most trajectories were produced
with the same agent harness, tool interface, stage definitions, artifact
templates, and Generator--Advisor protocol. This consistency improves
auditability, but it can introduce common stylistic and behavioral patterns.
An agent trained primarily on this corpus may therefore overfit to the current
harness and generalize poorly to different tool APIs, interaction protocols,
research domains, shorter workflows, or human--agent collaboration settings.
Moreover, 1,170 trajectories cover only a small fraction of the possible
research questions and workflows; long trajectories do not substitute for a
larger number of independent research processes.

Reverse synthesis introduces an additional source of bias. Although the final
proposal and Advisor-only constraints are hidden from the Generator, the
Advisor and anchor artifacts are still derived from a known endpoint. Generated
trajectories may therefore be more coherent and convergent than research in the
wild, underrepresent unsuccessful exploration, or inherit assumptions embedded
in the selected final artifact. Retaining idea-only trajectories avoids forcing
every partially shifted idea into a proposal, but it does not eliminate this
hindsight bias.

Finally, automated verification cannot fully establish scientific correctness,
naturalness, or downstream training utility. Search results and web content may
be incomplete or change over time; paper interpretations may remain shallow;
and selection based on human-curated artifacts and optional researcher
portraits may introduce domain, author, and venue biases. The current report
also does not establish through large-scale human evaluation or downstream
training experiments that value-tier weighting or the synthesized trajectories
improve research agents across settings.

We therefore view the main value of this work not as presenting a definitive
or exhaustive dataset, but as providing a concrete reference recipe for
synthesizing research-process supervision: start from high-quality final
artifacts, construct hidden Advisor constraints and intermediate anchors,
generate the visible tool-using trajectory forward, and verify its grounding
and causal order. Future releases should expand the number of trajectories,
models, harnesses, tools, and research domains; apply stronger post-processing
and human auditing; and evaluate transfer to independently implemented research
agents.

\end{document}